%% file: main.tex
\documentclass[10pt,journal,compsoc]{IEEEtran}
\ifCLASSINFOpdf
\else
\fi
\usepackage[utf8]{inputenc} 
\usepackage[T1]{fontenc}    

\usepackage{epsfig}
\usepackage{graphicx}
\usepackage{amsmath}
\usepackage{amssymb}
\usepackage{enumitem}

\usepackage{amsmath}    
\usepackage{bm}         
\usepackage{multirow}   
\usepackage{graphicx}   
\usepackage{booktabs}   
\usepackage{amssymb}    
\usepackage[edges]{forest}
\usepackage{xcolor}
\definecolor{hidden-draw}{RGB}{20,68,106}
\definecolor{hidden-pink}{RGB}{255,245,247}
\usepackage{tikz}

\usepackage[pagebackref,breaklinks,colorlinks]{hyperref}
\usepackage[capitalize]{cleveref}

\usepackage{xcolor}
\usepackage{tcolorbox}

\usepackage[linesnumbered,algo2e,boxed]{algorithm2e}
\usepackage[figuresright]{rotating}
\usepackage{amsmath,amssymb,textcomp,subfigure,multirow,upgreek}
\usepackage{booktabs}
\usepackage{color,mdwlist}
\usepackage{tikz}
\usepackage[edges]{forest}
\definecolor{hidden-draw}{RGB}{20,68,106}
\definecolor{hidden-pink}{RGB}{255,245,247}

\makeatletter
\DeclareRobustCommand\onedot{\futurelet\@let@token\@onedot}
\def\@onedot{\ifx\@let@token.\else.\null\fi\xspace}

\makeatother

\usepackage{ragged2e}

\usepackage{review}

\usepackage{graphicx,fontawesome}
\graphicspath{{figs/}}

\usepackage{caption}
\usepackage{wrapfig}

\usepackage{enumitem}
\setitemize{noitemsep,topsep=0pt,parsep=0pt,partopsep=0pt}

\begin{document}
\title{Aligning Multimodal LLM with Human Preference: A Survey}

\author{Tao Yu,
    Yi-Fan Zhang$\dagger$,
    Chaoyou Fu,
    Junkang Wu, 
    Jinda Lu,
    Kun Wang,
    Xingyu Lu,
    Yunhang Shen,\\
    Guibin Zhang,
    Dingjie Song,
    Yibo Yan,
    Tianlong Xu,
    Qingsong Wen,
    Zhang Zhang,
    Yan Huang,\\
    Liang Wang,~\IEEEmembership{Fellow,~IEEE}
    and Tieniu Tan,~\IEEEmembership{Fellow,~IEEE}
	\IEEEcompsocitemizethanks{
 \IEEEcompsocthanksitem
$\dagger$Yi-Fan Zhang is the project leader: yifanzhang.cs@gmail.com.
 \protect
	    \IEEEcompsocthanksitem 
     Tao Yu, Yi-Fan Zhang, Zhang Zhang, Yan Huang, Liang Wang, Tieniu Tan are with the Institute of automation, Chinese academy of science.
        Chaoyou Fu is with the Nanjing University.
        Junkang Wu, and Jinda Lu are with the University of Science and Technology of China.
        Kun wang is with the Nanyang Technological University.
        Xingyu Lu is with the Shenzhen International Graduate School, Tsinghua University.
        Yunhang Shen is with the Xiamen University.
        Guibin Zhang is with the National University of Singapore.
        Dingjie Song is with the Lehigh University.
        Yibo Yan is with the The Hong Kong University of Science and Technology.
        Tianlong Xu and Qingsong Wen are with the Squirrel Ai Learning.
        \protect
        }
}

%
%

\markboth{Journal of \LaTeX\ Class Files,
~October~2024}%
{Shell \MakeLowercase{\textit{et al.}}: Bare Demo of IEEEtran.cls for Computer Society Journals}
%



\IEEEtitleabstractindextext{%
\input{draft/000.abstract}

\begin{IEEEkeywords}
Multimodal Large Language Model, MLLM Alignment, Alignment with Human Preference.
\end{IEEEkeywords}}

\maketitle

\IEEEdisplaynontitleabstractindextext

%
\IEEEpeerreviewmaketitle


%
%
%
%


\input{draft/010_intro}

\input{draft/020.app_scenarios}
\input{draft/021.general_image}

\input{tables/loss_table}
\input{draft/022.multi-image}
\input{draft/023.extended}

\input{tables/data_table}
\input{draft/030.data}
\input{draft/031.external_knowledge}
\input{draft/032.self-annotation}
\input{draft/040.evaluation}
\input{draft/050.future_work}
\input{draft/060.conclusion}

\footnotesize
\bibliographystyle{IEEEtran}
\bibliography{main}

\end{document}

%% file: draft/000.abstract.tex
\justify
\begin{abstract}

Large language models (LLMs) can handle a wide variety of general tasks with simple prompts, without the need for task-specific training. Multimodal Large Language Models (MLLMs), built upon LLMs, have demonstrated impressive potential in tackling complex tasks involving visual, auditory, and textual data. However, critical issues related to truthfulness, safety, o1-like reasoning, and alignment with human preference remain insufficiently addressed. This gap has spurred the emergence of various alignment algorithms, each targeting different application scenarios and optimization goals. Recent studies have shown that alignment algorithms are a powerful approach to resolving the aforementioned challenges. In this paper, we aim to provide a comprehensive and systematic review of alignment algorithms for MLLMs. Specifically, we explore four key aspects: (1) the application scenarios covered by alignment algorithms, including general image understanding, multi-image, video, and audio, and extended multimodal applications; (2) the core factors in constructing alignment datasets, including data sources, model responses, and preference annotations; (3) the benchmarks used to evaluate alignment algorithms; and (4) a discussion of potential future directions for the development of alignment algorithms. This work seeks to help researchers organize current advancements in the field and inspire better alignment methods. The project page of this paper is available at \url{https://github.com/BradyFU/Awesome-Multimodal-Large-Language-Models/tree/Alignment}.
\end{abstract}

%% file: draft/010_intro.tex
\section{Introduction}

\IEEEPARstart{L}{LMs} have ushered in a new era for artificial intelligence (AI), demonstrating remarkable abilities such as instruction-following and few-shot learning~\cite{brown2020language}, which stem from their extensive model parameters and vast training data. These models represent a paradigm shift from traditional, task-specific models, as LLMs can handle a wide variety of general tasks with a simple prompt, without the need for task-specific training. This capability has fundamentally changed the AI landscape. However, while LLMs excel in text processing, they are limited by their inability to process multimodal data. Our world, on the other hand, is inherently multimodal, comprising visual, auditory, textual and other forms of data. This limitation has inspired the development of MLLMs~\cite{fu2024mme}, which extend LLMs by incorporating the ability to process and understand multimodal data. MLLMs open up new opportunities for applications that require the integration and understanding of multiple types of data, expanding the potential of AI.
\begin{figure*}[t]
\centering
\includegraphics[width=\linewidth]{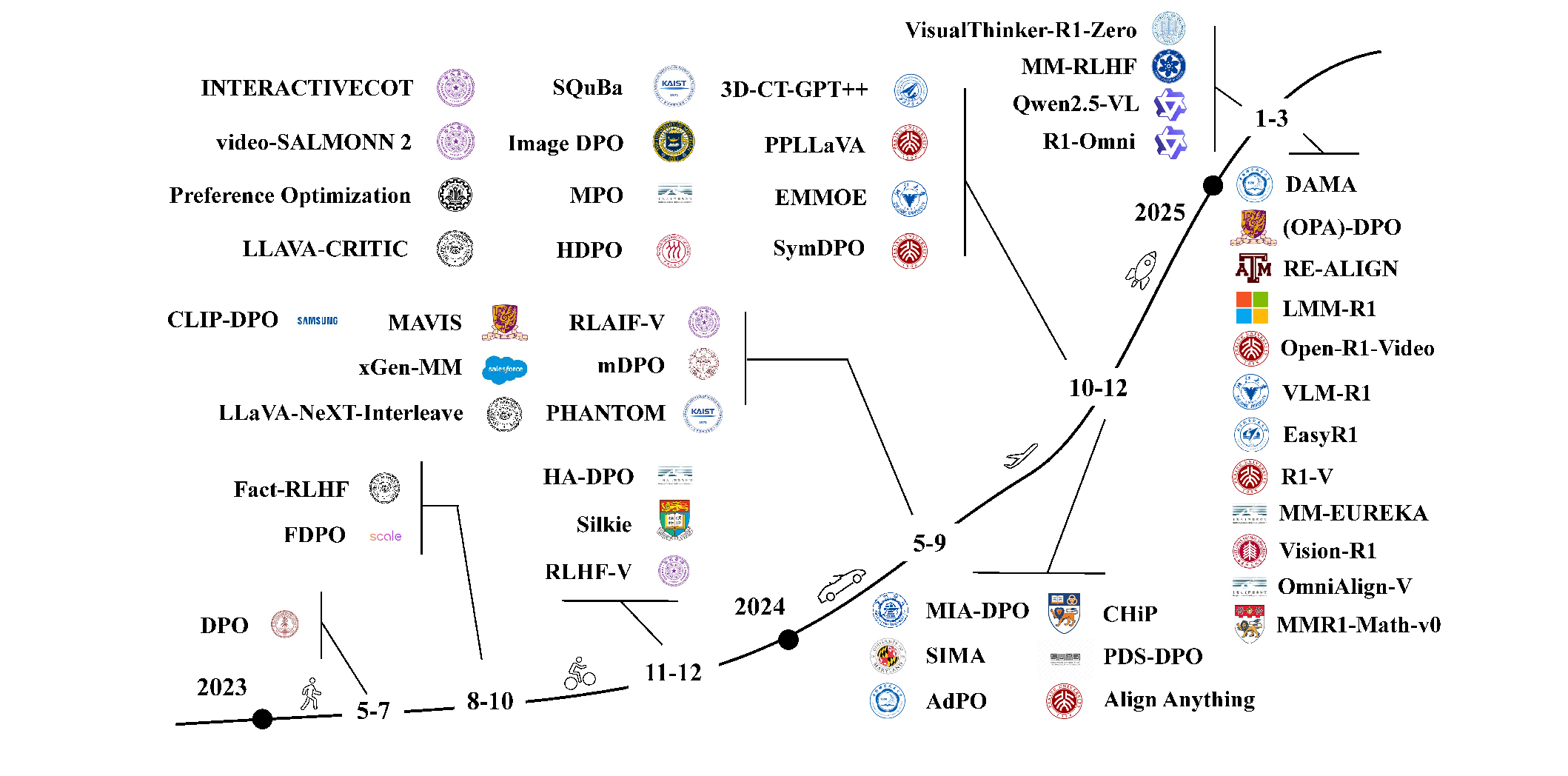}
\caption{ A timeline of MLLM alignment algorithms.}
\label{fig3}
\vspace{-0.3cm}
\end{figure*}

Despite the impressive potential demonstrated by MLLMs in tackling complex tasks that involve visual, auditory, and textual data, the current state-of-the-art MLLMs have rarely undergone rigorous alignment stage such as reinforcement learning from human preference (RLHF) stages~\cite{wang2024qwen2,deitke2024molmo,chen2024far,dai2024nvlm,agrawal2024pixtral,fu2025vita} and direct preference optimization (DPO \cite{rafailov2024directpreferenceoptimizationlanguage}). Typically, these models only advance to the supervised fine-tuning (SFT) phase, with critical issues related to authenticity, safety, and alignment with human preference remaining inadequately addressed. This gap has led to the emergence of various alignment algorithms, each targeting different application areas and optimization goals. However, this rapid development (Figure \ref{fig3}) also presents a number of challenges for researchers, particularly in areas such as benchmarking, optimizing alignment data, and introducing novel algorithms. In response, this paper provides a comprehensive and systematic review of alignment algorithms, focusing on the following four key questions:

\begin{itemize}
    \item \textbf{What application scenarios do existing alignment algorithms cover?} We categorize current alignment algorithms based on their application scenarios in Figure \ref{siweidaotu}, offering a clear framework for researchers across different domains. We also establish a unified symbolic system to aid researchers in understanding the distinctions and connections between various algorithms, which is summarized in Table \ref{tab:loss}.
    
    \item \textbf{How are alignment datasets constructed?} The creation of alignment datasets involves three core factors: data sources, model responses, and preference annotations. We perform a systematic analysis and categorization of these factors (publicly available datasets are summarized in Table \ref{tab:data}), highlighting the strengths and weaknesses of current construction methods and highlighting key considerations that need to be addressed.
    
    \item \textbf{How are alignment algorithms evaluated?} Given that most alignment algorithms are designed for specific tasks—such as addressing hallucinations, ensuring safety, and improving reasoning—we categorize and organize common alignment algorithm benchmarks, providing a clear framework for evaluation.
    
    \item \textbf{What are the future directions for the development of alignment algorithms? } We propose several potential future directions, such as the integration of visual information into alignment algorithms, insights from LLM alignment methods, and the challenges and opportunities posed by MLLMs as agents.
\end{itemize}

Although many existing surveys focus on the alignment of AI \cite{ji2024aialignmentcomprehensivesurvey,fu2024mme,kumar2025llmposttrainingdeepdive}, none of them specifically address the alignment of MLLMs. To the best of our knowledge, this survey is the first to specifically focus on the alignment of MLLMs. Our objective is to provide a comprehensive and systematic guide for researchers in both academia and industry, helping them identify appropriate tools and methodologies in the rapidly evolving field of MLLM alignment.

\input{tables/siweidaotu}

\section{Background: MLLM Alignment}
\begin{figure*}[t]
\centering
\includegraphics[width=0.9\linewidth]{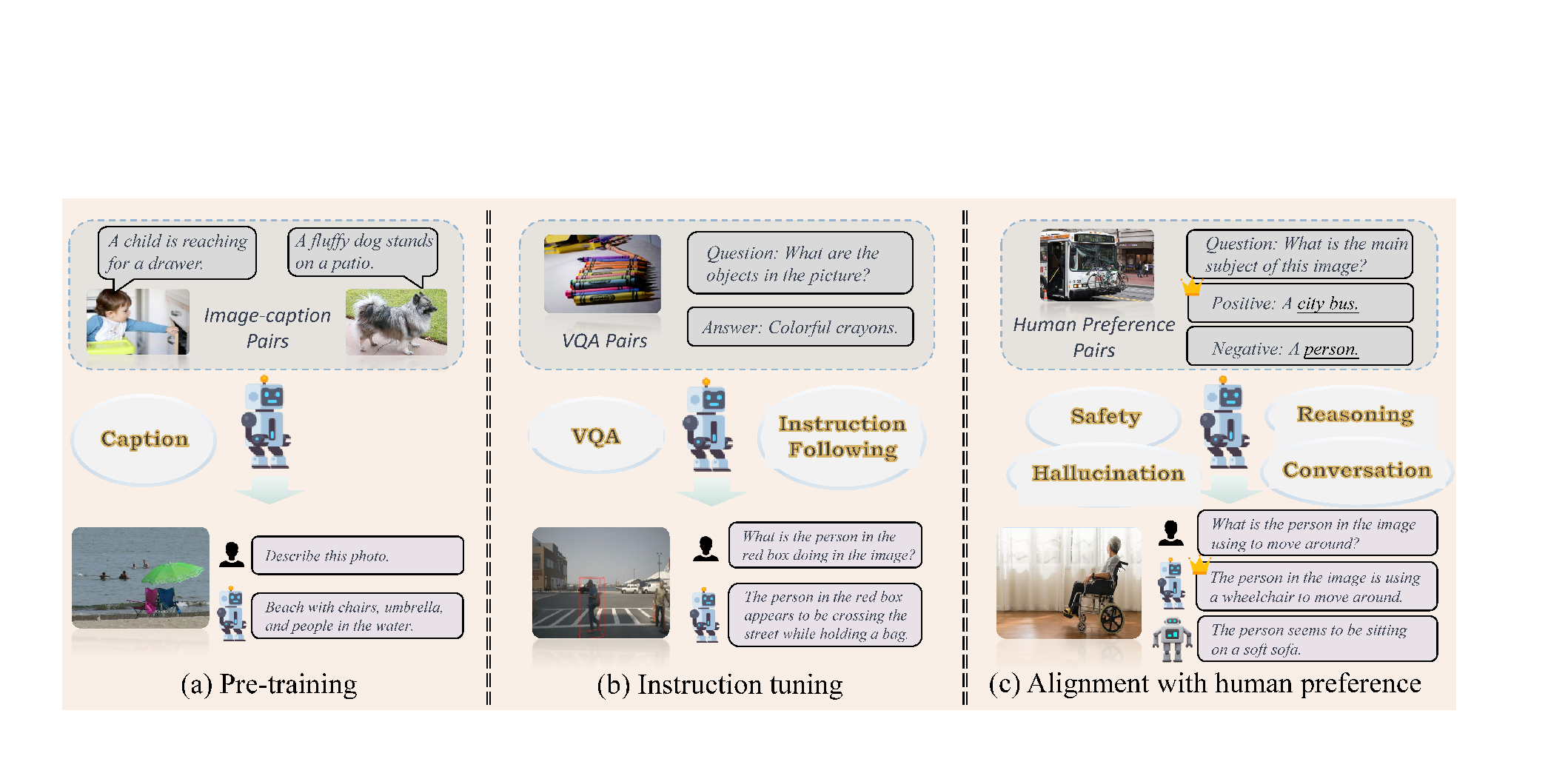}        
\caption{Comparison of pre-training, instruction tuning, and alignment with human preference.}
\label{fig1}
 
\end{figure*}

In this section, we will provide a brief explanation of the complete training process for MLLMs, which consists mainly of three phases (Figure \ref{fig1}): pre-training, instruction tuning, and alignment with human preference.

\textbf{Pre-Training.}
The pre-training phase of MLLMs primarily aims to align the feature spaces of different modalities with that of the language model. The data used in this phase is typically simple caption data. For instance, image-caption pairs are commonly used for image/video understanding MLLMs~\cite{bai2023qwen,chen2023internvl}, while speech data and transcriptions are used for speech understanding MLLMs~\cite{fu2024vita,fu2025vita}. Through this pre-training phase, the model learns to understand inputs from various modalities.

\textbf{Instruction Tuning.}
Building on the pre-training phase, the SFT phase aims to teach the model how to interact with humans by focusing on understanding questions and providing responses in a specified format, i.e., instruction-following ability. The data used in this phase is typically high-quality and diverse dialogue data. For example, in the commonly seen visual question answering (VQA) task, given an image and its corresponding instruction, the trained model will provide the correct answer for the task.

\textbf{Alignment with Human Preference.}
Previous works have shown that SFT tends to make the model memorize training data and try to generalize across diverse scenarios~\cite{chu2025sft}. The alignment phase, typically involving reinforcement learning (RL) strategies, is crucial for generalizing to unseen domains. However, most multimodal models neglect this step~\cite{wang2024qwen2,deitke2024molmo,chen2024far,dai2024nvlm,agrawal2024pixtral}. The goals of alignment stage are broad, such as reducing hallucinations~\cite{zhang2024debiasing,lu2025dama}, enhancing conversational abilities~\cite{xiong2024llavacriticlearningevaluatemultimodal}, improving safety~\cite{zong2024safety}, strengthening the reasoning abilities~\cite{wang2024enhancingreasoningabilitymultimodal}, improving capabilities for long-reasoning tasks like DeepSeek-R1~\cite{chen2025r1v}, and overall MLLM performance~\cite{zhang2025mmrlhfstepforwardmultimodal}.
This phase usually uses pair data that incorporates human preference.

%% file: tables/siweidaotu.tex
\tikzstyle{my-box}=[
    rectangle,
    draw=hidden-draw,
    rounded corners,
    text opacity=1,
    minimum height=1.5em,
    minimum width=5em,
    inner sep=2pt,
    align=center,
    fill opacity=.5,
    line width=0.8pt,
]
\tikzstyle{leaf}=[my-box, minimum height=1.5em,
    fill=hidden-pink!80, text=black, align=left,font=\footnotesize,
    inner xsep=2pt,
    inner ysep=4pt,
    line width=0.8pt,
]
\begin{figure*}[!th]
    \centering
    \resizebox{\textwidth}{!}{
        \begin{forest}
            forked edges,
            for tree={
                grow=east,
                reversed=true,
                anchor=base west,
                parent anchor=east,
                child anchor=west,
                base=center,
                font=\large,
                rectangle,
                draw=hidden-draw,
                rounded corners,
                align=left,
                minimum width=4em,
                edge+={darkgray, line width=1pt},
                s sep=3pt,
                inner xsep=2pt,
                inner ysep=3pt,
                line width=0.8pt,
                ver/.style={rotate=90, child anchor=north, parent anchor=south, anchor=center},
            },
            where level=1{text width=5.9em,align=center,fit=band,font=\normalsize,}{},
            where level=2{text width=8.5em,align=center,fit=band,font=\footnotesize,}{},
            where level=3{text width=8.1em,font=\tiny,}{},
            where level=4{text width=12.5em, font=\tiny,}{},
            [
                \ \ MLLM \\Alignment
                [
                    Application \\ Scenarios 
                    [
                        General Image \\ Understanding 
                            [   
                                    Mitigating Hallucinations
                                        [
                                            Fact-RLHF\cite{sun2023aligninglargemultimodalmodels}{, } DDPO\cite{yu2024rlhfvtrustworthymllmsbehavior}{, } FDPO\cite{gunjal2024detectingpreventinghallucinationslarge}{, }\\ HA-DPO\cite{zhao2024hallucinationsenhancinglvlmshallucinationaware}{, } mDPO\cite{wang2024mdpoconditionalpreferenceoptimization}{, } RLAIF-V\cite{yu2024rlaifvopensourceaifeedback}{, }\\ xGen-MM\cite{xue2024xgenmmblip3familyopen}{, } CHiP\cite{fu2025chipcrossmodalhierarchicaldirect}{, } HDPO\cite{fu2024mitigatinghallucinationmultimodallarge}{, }\\ DAMA\cite{lu2025dama}{, }
                                            (OPA)-DPO\cite{yang2025mitigatinghallucinationslargevisionlanguage}{, }
                                            RE-ALIGN \cite{xing2025realignaligningvisionlanguage}
                                        ]
                            ]
                            [
                                    Comprehensive Capabilities
                                        [
                                            MM-DPO\cite{zhang2025mmrlhfstepforwardmultimodal}{, }
                                            Silkie\cite{li2023silkiepreferencedistillationlarge}{, } 
                                            CLIP-DPO\cite{ouali2024clipdpovisionlanguagemodelssource}{, } \\SIMA\cite{wang2024enhancingvisuallanguagemodalityalignment}{, } 
                                            LLaVA-Critic\cite{xiong2024llavacriticlearningevaluatemultimodal}{, }        MPO\cite{wang2024enhancingreasoningabilitymultimodal}{, } \\
                                            Image DPO\cite{luo2025vvlm}{, }
                                            PDS-DPO \cite{wijaya2024multimodalpreferencedatasynthetic}{, } 
                                            Qwen2.5-VL \cite{bai2025qwen25vltechnicalreport}{, }\\
                                            OmniAlign-V \cite{zhao2025omnialignvenhancedalignmentmllms}{, } MM-RLHF~\cite{zhang2025mmrlhfstepforwardmultimodal}
                                        ]
                            ]
                            [
                                    Multi-Modal O1 Development
                                        [
                                            LMM-R1 \cite{peng2025lmmr1}{, }
                                            Open-R1-Video \cite{wang-2025-open-r1-video}{, } 
                                            VLM-R1 \cite{shen2025vlmr1}{, } \\EasyR1 \cite{zheng2025easyr1}{, } 
                                            R1-V \cite{chen2025r1v}{, }
                                            MM-EUREKA \cite{MM-EUREKA2025}{, }\\
                                            VisualThinker-R1-Zero \cite{zhou2025r1zerosahamomentvisual}{, }
                                            Vision-R1 \cite{huang2025visionr1incentivizingreasoningcapability}{, }\\
                                            R1-Omni\cite{zhao2025r1omniexplainableomnimultimodalemotion}{, }
                                            MMR1-Math-v0 \cite{MMR1-Math2025}
                                        ]
                            ]    
                    ]        
                    [
                        Multi-Image{,} Video{,} \\ and Audio 
                            [
                                Multi-Image
                                [
                                    MIA-DPO\cite{liu2024miadpomultiimageaugmenteddirect}
                                ]
                            ]
                            [
                                ICL
                                [
                                    SymDPO\cite{jia2024symdpoboostingincontextlearning}
                                ]
                            ]
                            [
                                Video
                                [
                                    LLaVA-NEXT-Interleave\cite{li2024llavanextinterleavetacklingmultiimagevideo}{, } 
                                    PPLLaVA\cite{liu2025ppllava}{, } \\
                                    MM-RLHF~\cite{zhang2025mmrlhfstepforwardmultimodal}
                                ]
                            ]
                            [
                                Audio-Visual
                                [
                                    Video-SALMONN 2\cite{tang2025enhancing}
                                ]
                            ]
                            [
                                Audio-Text
                                [
                                    SQuBa\cite{eom2025squba}
                                ]
                            ]    
                    ]      
                    [
                        Extended Multimodal \\ Applications
                            [
                                Medicine
                                    [
                                        3D-CT-GPT++\cite{chen2025dctgpt}
                                    ]
                            ]
                            [
                                Mathematics
                                    [
                                        MAVIS\cite{zhang2024mavismathematicalvisualinstruction}
                                    ]
                            ]
                            [
                                Embodied Intelligence
                                    [
                                        INTERACTIVECOT\cite{jiao2025interactivecot}{, } 
                                        EMMOE\cite{li2025homiebot}
                                    ]
                            ]
                            [
                                Safety
                                    [
                                        AdPO\cite{liu2024adpo}{, } VLGuard\cite{zong2024safetyfinetuningalmostcost}{, } 
                                        PO\cite{afzali2025aligning}{, } MM-RLHF~\cite{zhang2025mmrlhfstepforwardmultimodal}
                                    ]
                            ]
                            [
                                Agent
                                    [
                                        RL4VLM\cite{zhai2024finetuninglargevisionlanguagemodels}
                                    ]
                            ]
                    ]
                ]
                [                
                    Dataset \\ Construction
                    [
                        Using  External \\
                        \qquad Knowledge 
                        [
                             Human Annotation
                             [
                                LLaVA-RLHF\cite{sun2023aligninglargemultimodalmodels}{, } 
                                MM-RLHF\cite{zhang2025mmrlhfstepforwardmultimodal}{, }
                                RLHF-V\cite{yu2024rlhfvtrustworthymllmsbehavior}{, } \\
                                LLAMA 3.1\cite{grattafiori2024llama3herdmodels}{, } 
                                M-HalDetect\cite{gunjal2024detectingpreventinghallucinationslarge}
                             ]
                        ]
                        [
                            Closed-Source LLM/MLLM
                            [
                                LRV-Instruction\cite{liu2024mitigatinghallucinationlargemultimodal}{, } 
                                HA-DPO\cite{zhao2024hallucinationsenhancinglvlmshallucinationaware}{, }\\ 
                                Video-SALMONN 2\cite{tang2025enhancing}{, } 
                                PHANTOM\cite{lee2024phantomlatentlargelanguage}{, } \\
                                VLGuard\cite{zong2024safetyfinetuningalmostcost}{, } 
                                VLFeedback\cite{li2023silkiepreferencedistillationlarge}{, } OmniAlign-V \cite{zhao2025omnialignvenhancedalignmentmllms}{, }\\
                                MAVIS-Instruct\cite{zhang2024mavismathematicalvisualinstruction}{, } 
                                EMMOE-100\cite{li2025homiebot}{, } 
                                HDPO\cite{fu2024mitigatinghallucinationmultimodallarge}
                            ]
                        ]
                        [
                            Open-Source LLM/MLLM
                            [ 
                                LLaVA-Critic\cite{xiong2024llavacriticlearningevaluatemultimodal}{, }
                                INTERACTIVECOT\cite{jiao2025interactivecot}{, }\\
                                CLIP-DPO\cite{ouali2024clipdpovisionlanguagemodelssource}
                            ]
                        ]
                    ]
                    [
                        Self-Annotation
                            [
                                Single Text Modality
                                [
                                    SQuBa\cite{eom2025squba}{, } 
                                    SymDPO\cite{jia2024symdpoboostingincontextlearning}{, } 
                                    SIMA\cite{wang2024enhancingvisuallanguagemodalityalignment}{, }
                                    MMPR\cite{wang2024enhancingreasoningabilitymultimodal}{, } \\
                                    MIA-DPO\cite{liu2024miadpomultiimageaugmenteddirect}
                                ]
                            ]
                            [
                                Single Image Modality
                                [
                                    Image DPO\cite{luo2025vvlm}
                                ]
                            ]
                            [
                                Image-Text Mixed Modality
                                [
                                    AdPO\cite{liu2024adpo}
                                ]
                            ]
                    ]
                ]
                [
                    Evaluation \\ Benchmark
                        [
                               General Knowledge
                               [
                                    MME-RealWorld\cite{zhang2025mmerealworldmultimodalllmchallenge}{, } 
                                    MMStar\cite{chen2024rightwayevaluatinglarge}{, } 
                                    MMBench\cite{liu2023mmbench}{, } 
                                    MMT-Bench\cite{mmtbench}{, } 
                                    BLINK\cite{fu2024blink}{, } \\
                                    MathVista\cite{mathvista}{, } 
                                    SQA3D\cite{ma2022sqa3d}{, } 
                                    MMMU\cite{mmmu}{, }
                                    MVBench\cite{mvbench}{, } 
                                    Mantis-Instruct\cite{jiang2024mantisinterleavedmultiimageinstruction}
                                    , text width=22.2em
                               ]
                        ]
                        [
                               Hallucination
                                [
                                    Object HalBench\cite{rohrbach2019objecthallucinationimagecaptioning}{, } 
                                    VideoHallucer\cite{videohallucer}{, } 
                                    VALOR-Eval\cite{valor-eval}{, } 
                                    POPE\cite{pope}{, } 
                                    HaELM\cite{haelm}{, } \\
                                    OpenCHAIR\cite{openchair}{, } 
                                    GAVIE\cite{gavie}{, } 
                                    AMBER\cite{amber}{, } 
                                    Mementos\cite{mementos}{, } 
                                    MMHal-Bench\cite{mmhal-bench}{, }\\ 
                                    VLind-Bench\cite{vlind-bench}{, } 
                                    M-HalDetect\cite{gunjal2024detectingpreventinghallucinationslarge}{, } 
                                    HallusionBench\cite{hallusionbench}{, } 
                                    VHTest\cite{vhtest}{, } 
                                    RefoMB\cite{yu2024rlaifvopensourceaifeedback}{, }\\
                                    MHaluBench\cite{mhalubench}{, } 
                                    PhD\cite{phd}{, } 
                                    ActivityNet-QA\cite{activity-net-qa}{, } 
                                    R-Bench\cite{r-bench}{, } 
                                    MHumanEval\cite{raihan2025mhumanevalmultilingualbenchmark}{, } \\
                                    Bingo\cite{bingo}{, } 
                                    HQH\cite{hqh}
                                    , text width=22.2em
                                ]
                        ]
                        [
                                Safety
                                [
                                    AdvDiffVLM\cite{advdiffvlm}{, }
                                    RTVLM\cite{li2024redteamingvisuallanguage}{, }
                                    VLGuard\cite{zong2024safetyfinetuningalmostcost}{, }
                                    MultiTrust\cite{zhang2024multitrustcomprehensivebenchmarktrustworthy}{, }
                                    VLLM-safety-bench\cite{tu2023unicornsimagesafetyevaluation}{, }\\
                                    MOSSBench\cite{mossbench}{, }
                                    MM-RLHF-SafetyBench\cite{zhang2025mmrlhfstepforwardmultimodal}
                                    , text width=22.2em
                                ]
                        ]
                        [
                                Conversation
                                [
                                    Q-Bench\cite{q-bench}{, }
                                    LLVisionQA\cite{q-bench}{, }
                                    LLDescribe\cite{q-bench}{, }
                                    LLaVA-Bench-Wilder\cite{liu2024improvedbaselinesvisualinstruction}{, }\\
                                    LiveBench\cite{white2024livebenchchallengingcontaminationfreellm}{, }
                                    Vibe-Eval\cite{padlewski2024vibeevalhardevaluationsuite}
                                    , text width=22.2em
                                ]
                        ]
                        [
                                Reward Model
                                [
                                    M-RewardBench\cite{gureja2024mrewardbenchevaluatingrewardmodels}{, }
                                    VL-RewardBench\cite{li2024vlrewardbenchchallengingbenchmarkvisionlanguage}{, }
                                    RewardBench\cite{lambert2024rewardbenchevaluatingrewardmodels}{, }
                                    MJ-Bench\cite{chen2024mjbenchmultimodalrewardmodel}{, }\\
                                    MLLM-as-a-Judge\cite{chen2024mllmasajudgeassessingmultimodalllmasajudge}{, }
                                    MM-RLHF-RewardBench\cite{zhang2025mmrlhfstepforwardmultimodal}
                                    , text width=22.2em
                                ]
                        ]
                        [
                                Alignment
                                [
                                    Arena-Hard \cite{li2024crowdsourceddatahighqualitybenchmarks}{, }
                                    AlpacaEval-V2 \cite{dubois2024lengthcontrolledalpacaevalsimpleway}{, }
                                    AlignBench \cite{liu2024alignbenchbenchmarkingchinesealignment}{, }
                                    MM-AlignBench\cite{zhao2025omnialignvenhancedalignmentmllms}
                                    , text width=22.2em
                                ]
                        ]
                ]       
            ]
        \end{forest}
    }
    \caption{Categories of MLLM alignment, including application scenarios, alignment datasets, and evaluation benchmarks.}
    \label{siweidaotu}
\end{figure*}

%% file: draft/020.app_scenarios.tex
\section{Application Scenarios}

Recent advancements in MLLM alignment algorithms have significantly expanded their applicability across a variety of domains. As illustrated in Figure~\ref{siweidaotu}, these methods can be categorized into three tiers based on their application scenarios: (1) general image understanding; (2) alignment algorithms designed for more complex modalities (such as multi-image, video, and audio); and (3) extended applications targeting domain-specific tasks. The first tier establishes the foundational principles of MLLM alignment. The second tier addresses the challenges of integrating more diverse and complex modalities, enabling more comprehensive multimodal interactions. Finally, the third tier focuses on adapting alignment frameworks to meet the specialized requirements of specific applications. Together, these tiers represent a structured and progressive framework for advancing multimodal intelligence and broadening its practical impact. We unify all existing methods using consistent notations, as shown in Table~\ref{tab:loss}, making it easier for researchers to compare the differences and connections.


%% file: draft/021.general_image.tex
\subsection{General Image Understanding}

MLLM alignment algorithms are developed to address the issue of hallucinations in multimodal systems. Recent research shows that these algorithms not only improve performance in this regard but also enhance safety, conversational capabilities, reasoning abilities, and a range of other functional attributes. In this section, we systematically examine innovative approaches, categorizing them based on their primary application scenarios: mitigating hallucinations and enhancing additional capabilities.

\subsubsection{Mitigating Hallucinations } The original design intention of MLLM alignment algorithms is to mitigate hallucinations. 
Fact-RLHF \cite{sun2023aligninglargemultimodalmodels} is the first multimodal RLHF algorithm, utilizing 10K human-labeled samples for the reward model and 50K hold-out data. Its loss function integrates: (1) a per-token KL penalty; (2) factual information to calibrate judgments; and (3) correctness and length penalties. DDPO \cite{yu2024rlhfvtrustworthymllmsbehavior} assigns higher weights to corrected data in its loss function compared to standard DPO. It uses 1.4K manually refined samples covering hallucination types such as objects (41.2\%), positions (20.3\%), numbers (16.5\%), attributes (10\%), actions (5.3\%), and others (6.8\%). FDPO \cite{gunjal2024detectingpreventinghallucinationslarge} reuses InstructBLIP’s \cite{dai2023instructblipgeneralpurposevisionlanguagemodels} architecture by replacing the final embedding layer with a classification head to train clause/sentence-level reward models. After reward modeling and rejection sampling, it optimizes InstructBLIP with human-annotated data. HA-DPO \cite{zhao2024hallucinationsenhancinglvlmshallucinationaware} uses MLLM to generate image descriptions, validates them with GPT-4 for hallucinations, rewrites positive/negative samples for consistency, and adds an auxiliary causal language modeling loss to standard DPO. It reduces hallucinations and introduces the SHR metric (hallucinated sentences per total sentences). mDPO \cite{wang2024mdpoconditionalpreferenceoptimization} enhances DPO with a visual loss function (to counter visual information neglect) and anchoring (to prevent the decreasing in the probability of chosen response). RLAIF-V \cite{yu2024rlaifvopensourceaifeedback} generates multiple responses, splits each response into sentences, and reformulates each sentence as a question to query an open-source model for a trustworthiness score. The total score for each response is then used to determine the DPO data. xGen-MM \cite{xue2024xgenmmblip3familyopen} employs a four-stage pipeline (pre-training, SFT, interleaved multi-image supervised fine-tuning, post-training) to holistically improve hallucinations, helpfulness, and safety. CHiP \cite{fu2025chipcrossmodalhierarchicaldirect} introduces visual preference optimization (e.g., diffusion, cropping, rotation for negative images) and hierarchical text preferences (response/segment/token levels). Its loss combines visual DPO and three text-level DPO terms, targeting hallucination reduction. HDPO \cite{fu2024mitigatinghallucinationmultimodallarge} specifically constructs three hallucination-specific pairs: visual distracted hallucination, long context hallucination, and multimodal conflicts hallucination, aiming to reduce hallucinations.  DAMA \cite{lu2025dama} refines DPO with data hardness and model responses by adaptively modifying $\beta$. RE-ALIGN \cite{xing2025realignaligningvisionlanguage} intentionally injects controllable hallucinations into selected responses through image retrieval, generating rejected responses that provide more reasonable and natural preference signals regarding hallucinations. (OPA)-DPO \cite{yang2025mitigatinghallucinationslargevisionlanguage} prevents distribution shift by revising the responses generated by the pre-trained model using GPT-4V. The revised responses are then mixed with the ground truth to construct a dataset. The model is fine-tuned using LoRA-SFT \cite{hu2021loralowrankadaptationlarge}, converting out-of-distribution data into in-distribution data.

\subsubsection{Enhancing Comprehensive Capabilities } In this subsection, we introduce several algorithms designed to enhance various aspects of model performance beyond hallucination reduction. 
For instance, Silkie\cite{li2023silkiepreferencedistillationlarge} collects a diverse set of instruction datasets and correspondingly generates responses from 12 models, which are then evaluated using GPT-4V to obtain preference data for applying DPO. CLIP-DPO \cite{ouali2024clipdpovisionlanguagemodelssource} leverages CLIP \cite{radford2021learningtransferablevisualmodels} scores to label data and applies DPO loss, resulting in improvements in both hallucination mitigation and zero-shot classification tasks. SIMA \cite{wang2024enhancingvisuallanguagemodalityalignment} constructs preference pairs by having the model self-evaluate its own responses. LLaVA-Critic \cite{xiong2024llavacriticlearningevaluatemultimodal} uses LLaVA-OV\cite{li2024llavaonevisioneasyvisualtask} to generate responses, fine-tunes a critic model (LLaVA-Critic) for scoring, and iteratively applies DPO, thereby enhancing performance in hallucination reduction, image/video understanding, and open-ended dialogue. MPO \cite{wang2024enhancingreasoningabilitymultimodal} automates the construction of a diverse multimodal reasoning preference dataset and blends SFT loss with several preference optimization losses, leading to improvements in reasoning. Finally, Image DPO \cite{luo2025vvlm} perturbs images (e.g., via blurring or pixelation) while keeping textual inputs unchanged, optimizing performance through visual-only DPO loss. MM-DPO \cite{zhang2025mmrlhfstepforwardmultimodal} introduces a dynamic reward scale in DPO, where the reward model assigns higher weights to comparison pairs with larger reward margins during training. This ensures that the most informative samples have a greater impact on model updates. PDS-DPO \cite{wijaya2024multimodalpreferencedatasynthetic} uses synthetic images generated by Stable Diffusion \cite{rombach2022highresolutionimagesynthesislatent}, which are evaluated by a pre-trained reward model. The highest-rated images, along with preference data generated by an open-source MLLM and scored by Llama-3-8B-ArmoRM \cite{wang2024interpretablepreferencesmultiobjectivereward}, are used for DPO. This approach enhances the trustworthiness and reasoning capabilities of the MLLM. The DPO phase of Qwen2.5-VL \cite{bai2025qwen25vltechnicalreport} focuses on image-text data and pure text data, utilizing preference data to align the model with human preference, thereby enhancing the model's reasoning capabilities and task-specific performance. OmniAlign-V \cite{zhao2025omnialignvenhancedalignmentmllms} improves the alignment of MLLM with human preference by constructing a comprehensive dataset featuring diverse images, complex questions, and varied response formats.

\subsubsection{Multi-Modal O1 Development}
Recently, the popularity of DeepSeek-R1 \cite{deepseekai2025deepseekr1incentivizingreasoningcapability} has brought new inspiration to the MLLM community. LMM-R1 \cite{peng2025lmmr1} is trained on a pure-text math dataset using RLOO \cite{ahmadian2024basicsrevisitingreinforcestyle} and showed improvements on multimodal math benchmarks. Open-R1-Video \cite{wang-2025-open-r1-video} utilizes GRPO \cite{shao2024deepseekmathpushinglimitsmathematical} to enhance the model's performance in the video domain. VLM-R1 \cite{shen2025vlmr1} applies the R1 method to the referring expression comprehension (REC) task. EasyR1 \cite{zheng2025easyr1} proposed a multimodal RL training framework. R1-V \cite{chen2025r1v} attempts to improve the generalization capabilities of MLLMs. MM-EUREKA \cite{MM-EUREKA2025} has built a scalable multimodal large-scale reinforcement learning framework based on OpenRLHF \cite{hu2024openrlhf}, offering stronger scalability compared to R1-V. VisualThinker-R1-Zero \cite{zhou2025r1zerosahamomentvisual} successfully replicates R1's Aha Moment using only a 2B non-SFT model. Vision-R1 \cite{huang2025visionr1incentivizingreasoningcapability} is the first reasoning MLLM that combines cold-start and RL. R1-Omni \cite{zhao2025r1omniexplainableomnimultimodalemotion} focuses on the emotion recognition task and is the industry’s first application of reinforcement learning with verifiable reward (RLVR) with an omni-multimodal LLM. MMR1-Math-v0 \cite{MMR1-Math2025} remarkably achieves top-tier performance with just 6k high-quality samples from public training datasets.

Current advancements in optimizing MLLM alignment algorithms primarily focus on two critical dimensions: data and loss functions. In the realm of preference data collection, dominant strategies include manual annotation, strong model-generated data, and self-generation data. However, each of these approaches faces characteristic limitations. A persistent challenge lies in reducing annotation costs while simultaneously enhancing data quality and diversity. On the other hand, innovations in loss functions have introduced advanced variants of DPO, such as HDPO \cite{fu2024mitigatinghallucinationmultimodallarge} and DDPO \cite{yu2024rlhfvtrustworthymllmsbehavior}, which demonstrate significant potential. Additionally, frameworks like Image DPO \cite{luo2025vvlm} and CHiP \cite{fu2025chipcrossmodalhierarchicaldirect} incorporate vision-modality supervision, underscoring the importance of cross-modal alignment. Moving forward, progress in this field will hinge on two critical areas: improving data quality and diversity and optimizing multimodal loss functions to achieve more robust and efficient alignment.

%% file: draft/022.multi-image.tex
\subsection{Multi-Image, Video, and Audio}

Compared to single-image tasks, many natural scene tasks involve multiple images, videos, or audio, introducing not only richer contextual scenarios but also greater complexity. Addressing these challenges requires specialized architectural designs and domain-specific optimizations. For instance, multi-image tasks necessitate models capable of understanding the relationships between multiple inputs, while in-context learning (ICL) requires the extraction of relevant information from multiple contextually provided images. Similarly, video processing demands the ability to perceive and analyze a large sequence of frames, and the data format of audio streams differs significantly from visual modalities. To tackle these complexities, researchers are actively investigating novel architectural modifications and specialized training paradigms tailored to these tasks.

\begin{table*}[h]
\caption{Various preference optimization objectives given preference data $\mathcal{D} = (x, \mathcal{I}, y_{w}, y_{l}, r_{w}, r_{l})$, where x is the question, $\mathcal{I}$ is the Image, $y_{w}$ and $y_{l}$ are winning and losing responses, and $r_{w}$ and $r_{l}$ are there rewards scored by a reward model.}
\resizebox{\linewidth}{!}{%
\label{tab:loss}
\centering
\begin{tabular}{c | c}
\toprule
\textbf{Method} & \textbf{Loss} \\
\midrule
Fact-RLHF & 
   $\mathcal{L}_{\text{RLHF}}= 
- \bm{\mathrm{E}}_{(\mathcal{I},x) \in D,y \sim {\pi}_{\phi}(y|\mathcal{I},x)}[r_{\bm{\theta}}(\mathcal{I},x,y) - \beta \cdot \mathbb{D}_{KL}({\pi}_{\phi}(y|\mathcal{I},x)\parallel {\pi}^{\mathrm{INIT}}(y|\mathcal{I},x))] $ \\
\midrule
SILKIE,SIMA & \multirow{9}*{$\mathcal{L}_{\text{dpo}} = - \bm{\mathrm{E}}_{(\mathcal{I},x, y_{w}, y_{l}) \sim{\mathcal{D}}}[{\mathrm{log} \sigma} (\beta \mathrm{log} \frac{{\pi}_{\bm{\theta}}(y_{w}|\mathcal{I},x)}{{\pi}_{\mathrm{ref}}(y_{w}|\mathcal{I},x)}-\beta \mathrm{log} \frac{{\pi}_{\bm{\theta}}(y_{l}|\mathcal{I},x)}{{\pi}_{\mathrm{ref}}(y_{l}|\mathcal{I},x)})]$} \\
CLIP-DPO, RLAIF-V  & \\
3D-CT-GPT++, MAVIS & \\
EMMOE, xGen-MM(BLIP-3) & \\
LLaVA-NeXT-Interleave & \\
LLAVA-CRITIC & \\
SQuBa, PPLLaVA & \\
HDPO, SymDPO & \\
INTERACTIVECOT & \\
\midrule
\multirow{2}*{RLHF-V} & 
$\mathcal{L}_{\text{Dense-dpo}} = - \bm{\mathrm{E}}_{(\mathcal{I},x, y_{w}, y_{l})}[\mathbb{I}_{y_{i} \notin {y_{u}}}[{\mathrm{log} \sigma} (\beta \mathrm{log} \frac{{\pi}_{\bm{\theta}}(y_{w}|\mathcal{I},x)}{{\pi}_{\mathrm{ref}}(y_{w}|\mathcal{I},x)}-\beta \mathrm{log} \frac{{\pi}_{\bm{\theta}}(y_{l}|\mathcal{I},x)}{{\pi}_{\mathrm{ref}}(y_{l}|\mathcal{I},x)})] $\\
& 
$ + \mathbb{I}_{y_{i} \in {y_{u}}}[\gamma {\mathrm{log} \sigma} (\beta \mathrm{log} \frac{{\pi}_{\bm{\theta}}(y_{w}|\mathcal{I},x)}{{\pi}_{\mathrm{ref}}(y_{w}|\mathcal{I},x)}-\beta \mathrm{log} \frac{{\pi}_{\bm{\theta}}(y_{l}|\mathcal{I},x)}{{\pi}_{\mathrm{ref}}(y_{l}|\mathcal{I},x)})]]$
\\
\midrule
F-DPO & 
$\mathcal{L}_{\text{Fine grained-dpo}} = - \bm{\mathrm{E}}_{(\mathcal{I},x, y_{w}, y_{l})}[{\mathrm{log} \sigma} (\beta \mathrm{log} \frac{{\pi}_{\bm{\theta}}(y_{w}|\mathcal{I},x)}{{\pi}_{\mathrm{ref}}(y_{w}|\mathcal{I},x)}) - {\mathrm{log} \sigma} (\beta \mathrm{log} \frac{{\pi}_{\bm{\theta}}(y_{l}|\mathcal{I},x)}{{\pi}_{\mathrm{ref}}(y_{l}|\mathcal{I},x)})]$
\\
\midrule
HA-DPO & 
$\mathcal{L} = \mathcal{L}_{\text{dpo}} + \bm{\mathrm{E}}_{(\mathcal{I},x, y) \sim{\mathcal{D}_{\mathrm{SFT}}}}[- \mathrm{log P(y|\mathcal{I}, x; \pi_{\bm{\theta}})}]$
\\
\midrule
MIA-DPO & 
$\text{Loss} : \mathcal{L} = \mathcal{L}_{\text{dpo}} + \gamma \cdot \bm{\mathrm{E}}_{(\mathcal{I},x, y_{w}, y_{l}) \sim \mathcal{D}} [- \mathrm{log(y_{w}|\mathcal{I},x)}]$
\\
\midrule
\multirow{2}*{CHiP} & 
$\mathcal{L} = \mathcal{L}_{\text{dpo}} + \mathcal{L}_{\text{visual-dpo}} + \lambda \cdot \mathcal{L}_{\text{sentence-dpo}}
+ \gamma \cdot \bm{\mathrm{E}}_{(\mathcal{I}, x, y_{w}^{\mathrm{Token}}, y_{l}^{\mathrm{Token}})\sim{\mathcal{D}_{\mathrm{Token}}}} 
$ \\ & 
$
\beta \mathbb{D}_{\mathrm{SeqKL}}\left [ {\pi}_{\mathrm{ref}}(y_w|\mathcal{I},x)\parallel {\pi}_{\bm{\theta}}(y_w|\mathcal{I},x) \right ] - \beta \mathbb{D}_{\mathrm{SeqKL}} \left [{\pi}_{\mathrm{ref}}(y_l|\mathcal{I},x)\parallel {\pi}_{\bm{\theta}}(y_l|\mathcal{I},x)\right ]$
\\
\midrule
Image DPO & 
$\mathcal{L}_{\text{Image dpo}} = - \bm{\mathrm{E}}_{(\mathcal{I}_{w}, \mathcal{I}_{l}, x, y_{w})}[{\mathrm{log} \sigma} (\beta \mathrm{log} \frac{{\pi}_{\bm{\theta}}(y_{w}|\mathcal{I}_{w},x)}{{\pi}_{\mathrm{ref}}(y_{w}|\mathcal{I}_{w},x)}-\beta \mathrm{log} \frac{{\pi}_{\bm{\theta}}(y_{w}|\mathcal{I}_{l},x)}{{\pi}_{\mathrm{ref}}(y_{w}|\mathcal{I}_{l},x)})]$
\\
\midrule
\multirow{2}*{AdPO} & 
$\mathcal{L} = - \bm{\mathrm{E}}_{(\mathcal{I}_{w}, \mathcal{I}_{l}, x, y_{w}, y_{l})}[{\mathrm{log} \sigma} (\beta \mathrm{log} \frac{{\pi}_{\bm{\theta}}(y_{w}|\mathcal{I}_{w},x)}{{\pi}_{\mathrm{ref}}(y_{w}|\mathcal{I}_{w},x)}-\beta \mathrm{log} \frac{{\pi}_{\bm{\theta}}(y_{l}|\mathcal{I}_{l},x)}{{\pi}_{\mathrm{ref}}(y_{l}|\mathcal{I}_{l},x)})]$
\\
&
$ + \Sigma_{t=1}^{\textrm{T}}\textrm{log}{\pi}_{\bm{\theta}}(y_{w}^{t}|\mathcal{I}_{l},x_{t}^{1:t-1})$
\\
\midrule
PHANTOM & 
$\mathcal{L} = \mathcal{L}_{\mathrm{SFT}}- \bm{\mathrm{E}}_{(\mathcal{I}_{w}, \mathcal{I}_{l}, x, y_{w})}[{\mathrm{log} \sigma} (\frac{\beta}{|y_{w}|} \mathrm{log} {\pi}_{\bm{\theta}}(y_{w}|\mathcal{I}_{w},x)-(\frac{\beta}{|y_{w}|} \mathrm{log} {\pi}_{\bm{\theta}}(y_{w}|\mathcal{I}_{w},x))]$
\\
\midrule
video-SALMONN 2 & 
$\mathcal{L} = \mathcal{L}_{\text{dpo}} + \lambda \bm{\mathrm{E}}_{(\mathcal{I},x, y_{\textrm{gt}}) \sim{\mathcal{D}_{\mathrm{gt}}}} \mathrm{log}\pi_{\bm{\theta}}(y_{\mathrm{gt}}|\mathcal{I},x)$
\\
\midrule
Preference Optimization &
$\mathcal{L} = \mathcal{L}_{\text{dpo}} + \lambda \bm{\mathrm{E}}_{(\mathcal{I},x, y) \sim \mathcal{D}_{\mathrm{reg}}}[\mathrm{log} \mathrm{\frac{\pi_{\bm{\theta}}(y|x)}{\pi_{\mathrm{ref}}(y|x)}}]
$
\\
\midrule
DAMA & $ \mathcal{L} = - \bm{\mathrm{E}}_{(\mathcal{I},x, y_{w}, y_{l}) \sim{\mathcal{D}}}[{\mathrm{log} \sigma} (\alpha \cdot \beta \mathrm{log} \frac{{\pi}_{\bm{\theta}}(y_{w}|\mathcal{I},x)}{{\pi}_{\mathrm{ref}}(y_{w}|\mathcal{I},x)}- \alpha \cdot \beta \mathrm{log} \frac{{\pi}_{\bm{\theta}}(y_{l}|\mathcal{I},x)}{{\pi}_{\mathrm{ref}}(y_{l}|\mathcal{I},x)})]$
\\
\midrule
\multirow{2}*{mDPO} & 
$\mathcal{L} = \mathcal{L}_{\text{dpo}} + \bm{\mathrm{E}}_{(\mathcal{I}_{w}, \mathcal{I}_{l}, x, y_{w}, y_{l}) \sim{\mathcal{D}}}{[-{\mathrm{log} \sigma} (\beta \mathrm{log} \frac{{\pi}_{\bm{\theta}}(y_{w}|\mathcal{I}_{w},x)}{{\pi}_{\mathrm{ref}}(y_{w}|\mathcal{I}_{w},x)}-\beta \mathrm{log} \frac{{\pi}_{\bm{\theta}}(y_{l}|\mathcal{I}_{l},x)}{{\pi}_{\mathrm{ref}}(y_{l}|\mathcal{I}_{l},x)})]}
$ \\
& 
$
- {\mathrm{log} \sigma}(\beta \mathrm{log} \frac{{\pi}_{\bm{\theta}}(y_{w}|\mathcal{I}_{w},x)}{{\pi}_{\mathrm{ref}}(y_{w}|\mathcal{I}_{w},x)} - \delta)
$
\\
\midrule
\multirow{2}*{MPO} & 
$\mathcal{L} = \alpha_{1} \cdot \mathcal{L}_{\text{dpo}} - \alpha_{2} \cdot \bm{\mathrm{E}}_{(\mathcal{I}, x, y_{w}, y_{l}) \sim{\mathcal{D}}}\left [
 {\mathrm{log} \sigma}(\beta \mathrm{log} \frac{{\pi}_{\bm{\theta}}(y_{w}|\mathcal{I},x)}{{\pi}_{\mathrm{ref}}(y_{w}|\mathcal{I},x)} - \delta) \right ]$
\\
& 
$
- \alpha_{2} \cdot \left [
{\mathrm{log} \sigma}(\beta \mathrm{log} \frac{{\pi}_{\bm{\theta}}(y_{l}|\mathcal{I},x)}{{\pi}_{\mathrm{ref}}(y_{l}|\mathcal{I},x)} - \delta) \right ]
- \alpha_{3} \cdot \left [ \frac{\mathrm{log}{\pi}_{\mathrm{ref}}(y_{w}|\mathcal{I},x)}{|y_w|} \right ]
$
\\ \midrule
MM-RLHF & $ \mathcal{L} = - \bm{\mathrm{E}}_{(\mathcal{I},x, y_{w}, y_{l}) \sim{\mathcal{D}}}\left[{\mathrm{log} \sigma} \left(\beta\left(r_w-r_l\right) \mathrm{log} \frac{{\pi}_{\bm{\theta}}(y_{w}|\mathcal{I},x)}{{\pi}_{\mathrm{ref}}(y_{w}|\mathcal{I},x)}- \beta\left(r_w-r_l\right) \mathrm{log} \frac{{\pi}_{\bm{\theta}}(y_{l}|\mathcal{I},x)}{{\pi}_{\mathrm{ref}}(y_{l}|\mathcal{I},x)}\right)\right]$
\\
\bottomrule
\end{tabular}
}
\end{table*}%
\subsubsection{Multi-Image } While existing open-source MLLMs perform well on single-image tasks, they often struggle with multi-image contextual understanding. MIA-DPO \cite{liu2024miadpomultiimageaugmenteddirect} addresses the challenges of hallucination in multi-image scenarios by  concatenating unrelated single-image data into sequential, grid, and picture-in-picture images and constructing preference data. Specifically, the method analyzes the model’s attention patterns across multiple images to assign scores and extract positive-negative pairs. This approach not only achieves state-of-the-art performance on multi-image benchmarks but also maintains robustness in single-image tasks.

\subsubsection{ICL } Recent advancements in ICL for LLMs have inspired adaptations in MLLMs, but these models often suffer from textual over-reliance, which leads to the neglect of visual information. To address this issue, SymDPO \cite{jia2024symdpoboostingincontextlearning} employs the few-shot idea by replacing original text answers in examples with unrelated words. This modification reduces the influence of information provided by the text modality, encouraging the model to rely more on visual information for answers, thereby successfully improving performance on tasks such as image captioning and VQA.

\subsubsection{Video } Video understanding introduces greater risks of hallucinations compared to image-based tasks due to the added complexity of temporal dynamics. However, DPO-based alignment methods have demonstrated effectiveness in mitigating these errors. Current advancements adopt two strategic pathways: interleaved visual instruction tuning (e.g., LLaVA-NeXT-Interleave \cite{li2024llavanextinterleavetacklingmultiimagevideo}), which enhances multi-frame reasoning by combining interleaved visual instructions with DPO loss; granular video-text alignment (e.g., PPLLaVA \cite{liu2025ppllava}), employing fine-grained vision-prompt alignment, context length expansion via asymmetric positional encoding, and DPO optimization. These frameworks advance the performance of MLLMs on video tasks.

\subsubsection{Audio-Visual } While real-world videos typically contain audio, existing MLLMs lack audio processing capabilities. Video-SALMONN 2 \cite{tang2025enhancing} addresses audio modality blindness in MLLMs through a hierarchical framework: (1) audio-visual representation alignment via an audio aligner; (2) semantic fusion through joint audio-visual SFT; (3) generation optimization using multi-round reinforcement learning; and (4) capability restoration via "Rebirth" fine-tuning with self-generated high-quality data, enhancing audio-visual understanding capability in video analysis.

\subsubsection{Audio-Text } Abstract speech summarization struggles with redundancy in outputs. SQuBa \cite{eom2025squba} overcomes this through a three-phase framework: (1) aligning speech-text representations via ASR-focused projector training; (2) jointly fine-tuning LLM and projector; (3) using the SFT responses and answers generated by the fine-tuned model as pairs for DPO. This phased optimization synergizes speech understanding and conciseness while preserving inference efficiency.

The application of alignment algorithms in emerging multimodal domains is still in its early stages, highlighting two critical areas for exploration: designing task-specific data for novel fields and developing alignment algorithms that leverage the structural properties of specific modalities. 

%% file: draft/023.extended.tex
\subsection{Extended Multimodal Applications}


Most MLLMs are not originally designed with specific downstream tasks in mind, such as medical diagnostics, mathematical reasoning, embodied AI, safety-critical systems, and autonomous agents. However, their powerful multimodal processing capabilities have drawn significant interest from researchers and practitioners across various fields. Recently, several domain-specific alignment frameworks have been proposed to better adapt these models to downstream tasks. It is worth noting that these domain-specific applications exhibit substantial gaps compared to general image understanding tasks, necessitating specialized alignment paradigms to address their unique operational constraints and ethical considerations.


\subsubsection{Medicine } The deployment of MLLMs in clinical settings is often hindered by the high risk of erroneous medical diagnoses or other domain-specific errors. The 3D-CT-GPT++ framework \cite{chen2025dctgpt} addresses this issue through a DPO-based approach, utilizing GPT-4 \cite{gpt4} to score SFT model-generated medical reports and construct preference datasets for alignment. This human-free method significantly reduces diagnostic misalignments while achieving clinical-grade accuracy and coherence in AI-assisted imaging analysis.

\subsubsection{Mathematics } MLLMs struggle with math-vision integration due to dual challenges: insufficient domain-optimized training frameworks and fragile chain-of-thought (CoT) reasoning where minor errors trigger cascading solution failures. MAVIS \cite{zhang2024mavismathematicalvisualinstruction} addresses challenges in multimodal mathematical reasoning by enhancing MLLMs through a four-phase framework: (1) fine-tuning a math-specialized vision encoder through contrastive learning; (2) align the encoder with LLM; (3) instruction tuning strengthens step-by-step reasoning; (4) DPO refines logical coherence by aligning annotated CoT paths. This integrated approach achieves high performance in visual mathematical problem-solving benchmarks.

\subsubsection{Embodied Intelligence } Embodied intelligence research leverages MLLMs to advance agents’ reasoning through CoT optimization and hierarchical task decomposition. INTERACTIVECOT \cite{jiao2025interactivecot} enhances contextual reasoning via dynamic CoT optimization with domain-specific fine-tuning and real-time interaction feedback, boosting task success; EMMOE \cite{li2025homiebot} decomposes complex tasks into 966 subtasks, leveraging GPT-4 to create semantic-augmented datasets that improve embodied metrics like path efficiency. Together, they demonstrate how adaptive reasoning architectures and structured multimodal data engineering bridge the gap between semantic interpretation and actionable decision-making in embodied AI.

\subsubsection{Safety } The advancement of MLLMs introduces adversarial risks (e.g., harmful hallucination generation), several works propose their own solutions. AdPO \cite{liu2024adpo} strengthens robustness through contrastive DPO training on perturbed images, enhancing the resistance to attacks. VLGuard \cite{zong2024safetyfinetuningalmostcost} curates multimodal harmful content datasets and employs post-hoc fine-tuning to suppress unsafe behavior. In contrast, Preference Optimization (PO) \cite{afzali2025aligning} frames contrastive learning as a one-step Markov decision process, combining preference data for discrimination and regularization data for stability, primarily boosting robustness. These methods synergize adversarial resilience and safety alignment to address evolving security threats. MM-RLHF~\cite{zhang2025mmrlhfstepforwardmultimodal} artificially constructs datasets related to adversarial attacks, privacy, and security, and mix them with a large amount of general capability data. This approach simultaneously enhances both the model's security and its general capabilities, revealing that there is no strict trade-off between the two.

\subsubsection{Agent } The application of MLLMs in multi-step interactive decision-making is often limited, preventing their direct application in complex decision-making scenarios. To address this limitation, existing work \cite{zhai2024finetuninglargevisionlanguagemodels} introduces a proximal policy optimization (PPO \cite{schulman2017proximalpolicyoptimizationalgorithms})-driven alignment framework designed to optimize MLLMs for multi-round interactive decision-making. This approach effectively bridges the gap between semantic comprehension and actionable agent behaviors in dynamic, real-world scenarios.

The development of domain-specialized MLLMs will likely be driven by a synergistic co-evolution of alignment frameworks and domain-specific expertise. By tailoring alignment architectures to leverage the unique attributes and constraints of specific domains (e.g., healthcare, robotics, mathematics, and agent), these multimodal models can get greater effectiveness and precision.

%% file: tables/data_table.tex
\begin{table*}[t]
\caption{MLLM alignment datasets, including data size, categories, data sources, response model: the model to generate responses for training by given image and prompt, and annotation model: the model to annotate the responses.}
\scalebox{1.0}{
\label{tab:data}
\centering
\begin{tabular}{c|c|c|c|c|c}
\toprule
\textbf{Dataset} & \textbf{Size} & \textbf{Categories} & \textbf{Response Model} & \textbf{Data Sources} & \textbf{Annotation Model} \\
\midrule
LLaVA-RLHF & 10K & Hallucination & LLaVA-SFT & LLaVA-Instruct & Human \\
\midrule
RLHF-V & 1.4K & Hallucination & Muffin & UniMM-Chat & Human\\
\midrule
VLFeedback & 80K & Hallucination & 12 Models & 9 Datasets & GPT-4 \\
\midrule
CLIP-DPO & 750K & Hallucination & MobileVLM-v2 & 12 Datasets & CLIP \\
\midrule
M-HalDetect & 16K & Hallucination & InstructBLIP & MS COCO & Human \\
\midrule
HA-DPO & 6K & Hallucination & 3 Models & Visual Genome & GPT-4 \\
\midrule
SIMA & 17K & Hallucination & LLaVA-1.5 & LLaVA-Instruct & LLaVA-1.5 \\
\midrule
RLAIF-V & 83K & Hallucination & 3 Models & 7 Datasets & 2 Models \\ 
\midrule
xGen-MM (BLIP-3) & 62.6K & Hallucination & xGen-MM-4B & open-source & - \\
\midrule
MIA-DPO & 52K & Multi-Image & LLaVa-v1.5 \& InternLM-XC 2.5 & Not mentioned & Not mentioned \\
\midrule
MAVIS & 88K & Math & MAVIS-7B & Self-constructed & GPT-4 \\
\midrule
EMMOE-100 & 10K & Embodied AI & Video-LLaVA & Self-constructed & GPT-4 \\
\midrule
Image-DPO & 60K & visual reasoning & Cambrian-8B \& LLaVA-1.5 & 3 Datasets & Stable Diffusion \\
\midrule
LLAVA-CRITIC & 40.1K & Multiple tasks & LLaVA-OneVision & 3 Datasets &  LLaVA-OneVision \\
\midrule
MMPR & 3.25M & Reasoning & InternVL2-8B & Not mentioned & automate pipeline \\ \midrule
MM-RLHF & 120K & \begin{tabular}[c]{@{}c@{}}Hallucination\\ Math, Video, Safety \\ Conversation \end{tabular} & \begin{tabular}[c]{@{}c@{}}GPT-4o, QwenVL2-72B, \\LLaVA-Video-72B, LLaVA-ov-72B\\ Claude3.5-Sonnet\end{tabular} & \begin{tabular}[c]{@{}c@{}}7 Dataset \&\\ Self-constrcuted\end{tabular} & Human \\
\midrule
Open-R1-Video-4k & 4K & o1-Reasoning & GPT-4o & LLaVA-Video-178K & Not mentioned \\
\midrule
MM-Eureka-Dataset & 54K & o1-Reasoning & InternVL2.5-8B-instruct & 15 Datasets & Not mentioned \\
\midrule
MMR1-Math-RL-Data-v0 & 7K & o1-Reasoning & Not mentioned & Not mentioned & Not mentioned \\
\bottomrule
\end{tabular}
}
\end{table*}

%% file: draft/030.data.tex
\section{MLLM Alignment Dataset}

In this section, we classify existing MLLM alignment datasets into two categories based on their construction approach: datasets that introduce external knowledge and those that rely on self-annotation. Table~\ref{tab:data} presents crucial information about publicly available datasets, including data sources, response generation methods, annotation techniques, and dataset sizes, providing a convenient reference.

%% file: draft/031.external_knowledge.tex
\subsection{Introducing External Knowledge}

Introducing high-quality external knowledge during data construction can enhance the quality of the generated alignment data. However, balancing data quality, quantity, and cost is a key consideration. Several works have explored data construction based on external knowledge.


\subsubsection{Human Annotation } Multiple datasets employ distinct human annotation strategies for training: LLaVA-RLHF \cite{sun2023aligninglargemultimodalmodels} collects 10k examples by having annotators select positive/negative responses from model-generated pairs. RLHF-V \cite{yu2024rlhfvtrustworthymllmsbehavior} creates 1.4k positive examples by manually correcting hallucinated responses. LLAMA 3.1 \cite{grattafiori2024llama3herdmodels} incorporates 7-point ratings and optional human edits for "chosen" responses from a model pool. M-HalDetect \cite{gunjal2024detectingpreventinghallucinationslarge} introduces clause-level hallucination analysis (16k examples) to synthesize preference data but remains in the exploratory stage. MM-RLHF \cite{zhang2025mmrlhfstepforwardmultimodal} covers three domains: image, video understanding, and MLLM safety. Through a rigorous pipeline construction, MM-RLHF ensures high-quality, fine-grained human annotations.



\subsubsection{Closed-Source LLM/MLLM } As the best-performing MLLMs currently available, GPT-4 series models have achieved near-human accuracy across many tasks. To reduce costs, current methods use them for preference data construction. LRV-Instruction \cite{liu2024mitigatinghallucinationlargemultimodal} uses GPT-4 to generate a large, diverse dataset of 400k visual instructions covering 16 vision and language tasks. The dataset includes both positive and negative data, but the positive and negative data are generated separately. HA-DPO~\cite{zhao2024hallucinationsenhancinglvlmshallucinationaware} uses GPT-4 to modify negative responses from MLLMs into positive ones, and then has both the positive and negative responses further corrected by GPT-4 to ensure that the positive and negative examples remain within the same distribution. This method collects 10,000 data annotated by GPT-4. Video-SALMONN 2~\cite{tang2025enhancing} employs GPT-3.5/4o and Gemini-1.5-Pro \cite{geminiteam2024gemini15unlockingmultimodal} for caption generation. PHANTOM \cite{lee2024phantomlatentlargelanguage} extracts 2.8 million visual instruction-tuning data from multiple datasets, using GPT-4o-mini to generate ambiguous responses for queries as negative examples and filtering them with GPT-4o to improve data quality. PHANTOM's approach of generating ambiguous responses to obtain negative data is novel, but its effectiveness remains to be discussed. Task-specific datasets include VLFeedback \cite{li2023silkiepreferencedistillationlarge} (80k GPT-4V-scored responses across 12 MLLMs), MAVIS-Instruct \cite{zhang2024mavismathematicalvisualinstruction} (math CoT preference data), and EMMOE-100 \cite{li2025homiebot} (3.7k SFT data and 10k DPO data of embodied AI). OmniAlign-V \cite{zhao2025omnialignvenhancedalignmentmllms} filters images based on image complexity and the number of meaningful objects in the images, and generates positive pairs for DPO using GPT-4o.

\subsubsection{Open-Source LLM/MLLM} Considering the invocation cost of GPT-4 series models in constructing large-scale alignment data, current methods use open-source models for preference data construction. INTERACTIVECOT~\cite{jiao2025interactivecot} builds an agent in ALFWorld\cite{shridhar2021alfworldaligningtextembodied} using predefined scores for embodied intelligence preference datasets. CLIP-DPO~\cite{ouali2024clipdpovisionlanguagemodelssource} argues that scoring based on MLLMs lacks stable evaluation metrics, so it substitutes CLIP \cite{radford2021learningtransferablevisualmodels} scores with clear meanings to select DPO pairs and constructs a 750k dataset (mixed QA/caption pairs).

Overall, manual annotation ensures high-quality, preference-aligned data but is constrained by challenges such as subjectivity and high costs. Both closed-source models (e.g., GPT-4V) and open-source models reduce costs and enable the large-scale construction of datasets; however, they often compromise on data quality. Looking ahead, we look forward to the development of more efficient methods can achieve a balance between scalability and reliability.

%% file: draft/032.self-annotation.tex
\subsection{Self-Annotation}

Data generated with the assistance of humans or models like GPT-4 may exhibit significant distributional differences from the target model, leading to issues such as overlooking image details~\cite{zhou2024aligningmodalitiesvisionlarge}.
As a result, several approaches have emerged that do not rely on external models for data generation or reward signals, instead depending on the target model itself to construct preference pairs. Based on the modality differences in preference pair data, we categorize them into three types: single-text modality (where preference pairs differ only in the text modality), single-image modality (where preference pairs differ only in the image modality), and image-text mixed modality (where preference pairs differ in both modalities).

\subsubsection{Single Text Modality } SQuBa~\cite{eom2025squba} uses SFT data as questions and positive samples, and employs the responses generated by the fine-tuned model as negative samples for DPO. SymDPO~\cite{jia2024symdpoboostingincontextlearning} reorganizes VQA/classification data into ICL format with meaningless text symbols to enhance visual learning and select DPO pairs. SIMA~\cite{wang2024enhancingvisuallanguagemodalityalignment} avoids the use of third-party data and models by having the model evaluate its own generated responses to rank the answers. MMPR \cite{wang2024enhancingreasoningabilitymultimodal} uses the model's responses generated based on images as positive examples, and truncates these positive examples to create negative samples by continuing the response without providing the image. MIA-DPO~\cite{liu2024miadpomultiimageaugmenteddirect} concatenates single-image data into multi-image formats and selects preferences via attention values, improving multi-image task performance.


\subsubsection{Single Image Modality } Image DPO~\cite{luo2025vvlm} constructs DPO preference pairs by perturbing images (e.g., gaussian blur, or pixelation) while keeping text unchanged, creating negative examples.

\subsubsection{Image-Text Mixed Modality } AdPO~\cite{liu2024adpo} aligns adversarial training with DPO by constructing preference pairs from original/adversarial images (generated via methods like PGD \cite{madry2019deeplearningmodelsresistant}) and their model responses, where both images and texts differ between positive and negative examples during optimization.

The construction of self-annotated positive and negative samples helps mitigate distribution shifts. However, due to performance limitations of MLLMs, current data quality remains relatively low. We look forward to future developments will introduce technologies such as automated data enhancement specifically designed for self-annotation approaches to improve data quality.

%% file: draft/040.evaluation.tex
\section{Evaluation}

Existing MLLM alignment evaluation benchmarks are categorized into six key dimensions: general knowledge (assessing foundational capabilities), hallucination (measuring the inconsistency of generated content with facts), safety (evaluating the ability to mitigate risks in responses), conversation (testing whether the model can output the content required by users), reward model (evaluating the performance of the reward model), and alignment with human preference.

\subsection{General Knowledge}


Most benchmarks prioritize high-quality, human-annotated datasets tailored for real-world applications. Examples include MME-RealWorld’s \cite{zhang2025mmerealworldmultimodalllmchallenge} 29K QA pairs from 13K images and MMMU’s \cite{mmmu} 11.5K questions from academic sources. MMStar \cite{chen2024rightwayevaluatinglarge} enhances reliability by minimizing data leakage and emphasizing visual dependency. Many benchmarks introduce novel methodologies, such as MMBench’s \cite{liu2023mmbench} bilingual evaluation with CircularEval, MMT-Bench’s \cite{mmtbench} task graphs for in/out-of-domain analysis, and BLINK’s \cite{fu2024blink} focus on visual perception tasks. These frameworks enhance evaluation precision and reveal model limitations. Tasks often require advanced multimodal reasoning, such as MathVista’s \cite{mathvista} mathematical-visual integration, SQA3D’s \cite{ma2022sqa3d} 3D situational QA, and MMMU’s coverage of charts, and maps. These benchmarks push models to handle interdisciplinary challenges. By curating challenging, fine-grained tasks (e.g., temporal understanding in MVBench \cite{mvbench}, multi-image processing in Mantis-Instruct \cite{jiang2024mantisinterleavedmultiimageinstruction}), these benchmarks aim to advance models’ ability to solve real-world problems requiring nuanced perception and reasoning.


\subsection{Hallucination}


These benchmarks systematically identify and categorize hallucinations in multimodal models, including object hallucinations (Object HalBench \cite{rohrbach2019objecthallucinationimagecaptioning}), intrinsic and extrinsic hallucinations (VideoHallucer \cite{videohallucer}), and associative biases (VALOR-Eval \cite{valor-eval}). They emphasize granular evaluation across visual, textual, and sequential contexts. Many propose novel frameworks, such as polling-based queries (POPE \cite{pope}), LLM-driven scoring (HaELM \cite{haelm}, RefoMB \cite{yu2024rlaifvopensourceaifeedback}), open-vocabulary detection (OpenCHAIR \cite{openchair}), annotation-free assessment (GAVIE \cite{gavie}), LLM-free pipelines (AMBER \cite{amber}), and GPT-4-assisted reasoning analysis (Mementos \cite{mementos}). They emphasize automated, scalable evaluation while addressing limitations like data leakage (MMHal-Bench \cite{mmhal-bench}) and language priors (VLind-Bench \cite{vlind-bench}). Datasets prioritize fine-grained human annotations (M-HalDetect \cite{gunjal2024detectingpreventinghallucinationslarge}, HallusionBench \cite{hallusionbench}) and synthetic data generation (VHTest \cite{vhtest}, MHaluBench \cite{mhalubench}). They balance real-world complexity (PhD’s \cite{phd} counter-commonsense images, ActivityNet-QA’s \cite{activity-net-qa} 58K QA pairs) and controlled challenges (R-Bench’s \cite{r-bench} robustness analysis). Some target specialized tasks like multilingual support (MHumanEval \cite{raihan2025mhumanevalmultilingualbenchmark}), while others address broad issues like bias and interference (Bingo \cite{bingo}). All aim to enhance model robustness in practical scenarios. By proposing alignment strategies (RLAIF-V’s \cite{yu2024rlaifvopensourceaifeedback} open-source feedback) and proposing unified framework (HQH \cite{hqh}), these benchmarks guide the development of more reliable multimodal systems.

\subsection{Safety}

Some studys introduce novel techniques, such as diffusion-based adversarial attacks (AdvDiffVLM \cite{advdiffvlm}), red teaming frameworks (RTVLM \cite{li2024redteamingvisuallanguage}), and post-hoc fine-tuning strategies (VLGuard \cite{zong2024safetyfinetuningalmostcost}). These approaches enhance evaluation rigor by simulating real-world threats or improving model resilience. Benchmarks like MultiTrust \cite{zhang2024multitrustcomprehensivebenchmarktrustworthy} and RTVLM unify trustworthiness assessment across multiple dimensions (e.g., truthfulness, fairness), while others target specific challenges like out of distribution (OOD) generalization (VLLM-safety-bench \cite{tu2023unicornsimagesafetyevaluation}) or oversensitivity (MOSSBench \cite{mossbench}). Together, they provide holistic insights into model limitations. MM-RLHF-SafetyBench~\cite{zhang2025mmrlhfstepforwardmultimodal} samples from existing datasets, further covering areas such as adversarial attacks, privacy, red team attacks, and harmful content detection.

\subsection{Conversation}

These benchmarks prioritize evaluating foundational visual skills, such as low-level perception ability (Q-Bench \cite{q-bench}, LLVisionQA \cite{q-bench}), description ability on low-level information (LLDescribe \cite{q-bench}), and quality assessment. They emphasize the model’s ability to interpret and articulate fine-grained visual information. Several benchmarks test generalization to challenging scenarios, including unconventional images (LLaVA Bench-Wilder \cite{liu2024improvedbaselinesvisualinstruction}), cross-domain tasks (LiveBench’s \cite{white2024livebenchchallengingcontaminationfreellm} math/news integration), and adversarial prompts (Vibe-Eval’s \cite{padlewski2024vibeevalhardevaluationsuite} high-difficulty questions). They reveal model adaptability beyond standard datasets.


\subsection{Reward Model}

Each benchmark targets specific evaluation dimensions, such as multilingual capabilities (23 languages in M-RewardBench \cite{gureja2024mrewardbenchevaluatingrewardmodels}), alignment/safety/bias (MJ-Bench \cite{chen2024mjbenchmultimodalrewardmodel}), leveraging human annotations to enhance explainability and the final model scoring capability (MM-RLHF-RewardBench \cite{zhang2025mmrlhfstepforwardmultimodal}), and ability of MLLMs in assisting judges across diverse modalities (MLLM-as-a-Judge’s \cite{chen2024mllmasajudgeassessingmultimodalllmasajudge} scoring vs. pairwise comparisons). These frameworks reveal model strengths and weaknesses in structured and OOD scenarios. High-quality datasets are curated through human-AI collaboration (VL-RewardBench’s \cite{li2024vlrewardbenchchallengingbenchmarkvisionlanguage} annotation pipeline) or structured triplet designs (RewardBench \cite{lambert2024rewardbenchevaluatingrewardmodels}). Tasks range from simple preference ranking to complex reasoning, pushing models to handle nuanced challenges like hallucination and ethical alignment.

\subsection{Alignment}
Some benchmarks investigate the ability of models to align with human preference. Arena-Hard \cite{li2024crowdsourceddatahighqualitybenchmarks} is a comprehensive, multi-dimensional benchmark designed to evaluate the alignment of Chinese LLMs. AlpacaEval-V2 \cite{dubois2024lengthcontrolledalpacaevalsimpleway} proposes a simple regression analysis method to control for length bias in self-assessment. Arena-Hard \cite{li2024crowdsourceddatahighqualitybenchmarks} increases the separation of model performance by three times and achieves a 98.6\% correlation with human preference rankings. MM-AlignBench\cite{zhao2025omnialignvenhancedalignmentmllms} is a manually annotated benchmark specifically designed to assess the alignment with human values.

Overall, for MLLM alignment algorithms, many current works focus on their ability to prevent models from generating hallucinations, while also exploring how to leverage alignment algorithms to enhance MLLMs' general knowledge and conversation capability, which is an important direction for the future. Some researchers treat unsafe responses as misaligned with human preferences, thereby applying MLLM alignment algorithms to address safety issues. The effectiveness of reward models in these frameworks, particularly their performance in guiding alignment, warrants further investigation. Additionally, benchmarks for alignment with human preference have also evolved from the LLM domain to the MLLM domain.

%% file: draft/050.future_work.tex
\section{Future Work and Open Challenges}

As MLLMs rapidly advance, aligning them with human preferences has become a central focus. However, several challenges persist. First, there is a scarcity of high-quality and diverse datasets. Second, many methods fail to effectively utilize visual information, often relying primarily on text for constructing positive and negative samples, and neglecting the full potential of multimodal data. Additionally, there is a lack of comprehensive evaluation standards, with current methods often being validated only on specific types of benchmarks, such as hallucination or dialogue tasks, which makes it difficult to assess their generalizability. Furthermore, by drawing on advancements in LLM post-training strategies and agent research, we can pinpoint limitations in existing MLLM alignment approaches. Overcoming these challenges is essential for developing more robust and comprehensive alignment methods.

\subsection{Data Challenges} The alignment of MLLMs faces two critical data-related challenges: data quality and coverage. First, the availability of high-quality MLLM alignment data is limited. Compared to LLMs, acquiring and annotating multimodal data is significantly more complex due to the inherent difficulties of handling multiple modalities. Second, existing datasets lack sufficient coverage of diverse multimodal tasks, such as optical character recognition, mathematical problems, and chart understanding, among others. Constructing a comprehensive dataset that addresses this wide array of tasks is an extremely challenging endeavor. To the best of our knowledge, there is currently no publicly available, fully human-annotated multimodal dataset that exceeds 200,000 samples. These limitations in data quality and coverage pose significant barriers to effectively aligning MLLMs.

\subsection{Leveraging Visual Information for Alignment} For clarity, we use the following notation to represent the composition of current alignment data: preference data $\mathcal{D} = (x, \mathcal{I}, y_{w}, y_{l})$, where $x$ is the question, $\mathcal{I}$ is the image, and $y_{w}$ and $y_{l}$ represent the winning and losing responses, respectively. In current research, three main approaches are employed to leverage visual information in order to enhance alignment performance, though each has its limitations:

\begin{enumerate}
    \item \textbf{Using corrupted or irrelevant images as alignment phase negative samples.} Researchers create new images $\mathcal{I}_{neg}$ and use $(y_w | x, \mathcal{I}_{neg})$ as a negative sample. This approach improves MLLM robustness to different images and reduces hallucinations. However, visual negatives often rely on diffusion algorithms or image modifications that lack robust quality metrics, incurring high computational costs.
    
    \item \textbf{Generating new questions and answers based on corrupted images.} In this method, researchers create a new image $\mathcal{I}_{neg}$, use it to generate additional response $y_{neg}$, and then treat $(y_{neg} | x, \mathcal{I})$ as a negative sample. This method also essentially compares textual outputs, but it adds more variety to the textual comparison. However, the process of generating additional negative samples incurs extra computational overhead.
    
    \item \textbf{Using cosine similarity metrics from models like CLIP to assess text-image matching.} This approach uses a similarity score between the text and the image to filter data or as part of the reinforcement learning reward function. While this can help reduce data noise, the quality of the score depends on the evaluation model's quality, which may be subject to model bias.
\end{enumerate}

Each of these methods plays a role in enhancing MLLM alignment with visual data, but they come with trade-offs in terms of efficiency, cost, and the potential for biases.

\subsection{Comprehensive Evaluation } Most MLLM alignment studies primarily evaluate their algorithms on a few key areas, such as hallucination, conversational abilities, or safety. However, we argue that aligning MLLMs with human preference should not be restricted to these specific tasks. Future research should adopt a more comprehensive evaluation approach, assessing alignment methods across a broader range of tasks to better demonstrate their generalizability and effectiveness.

\subsection{Full-Modality Alignment } Align-anything\cite{ji2024alignanythingtrainingallmodality} pioneers full-modality alignment through the multimodal dataset "align-anything-200k", which spans text, images, audio, and video. This study demonstrates the complementary effects between different modalities. However, their work is still in its early stages. The dataset for each modality is relatively small, limiting its ability to cover a wide range of tasks. Additionally, the proposed algorithm is only a preliminary improvement on the DPO method, and it does not fully exploit the unique structural information inherent in each modality. Moving forward, the design of alignment algorithms beyond image/text domains, particularly for other modalities, to enhance multimodal model capabilities, will be a key trend.

\subsection{MLLM Reasoning } Recent reasoning LLMs represented by OpenAI (O1) \cite{gpt-o1} and DeepSeek-R1 \cite{deepseekai2025deepseekr1incentivizingreasoningcapability} have demonstrated that RL algorithms and preference data are crucial for improving LLMs in complex problem-solving, long-context understanding, and generation tasks. 
Here we will explore the insights gained from LLM reasoning enhancement research and their implications for aligning MLLMs from two dimensions: data and optimization framework.

(1) \textit{Data}. i) \textit{Scale \& quality}. Corresponding approaches have gradually evolved from the small-model resampling (e.g., OpenMathInstruct\cite{toshniwal2024openmathinstruct118millionmath}) to high-quality synthetic data \cite{gpt4} (e.g., AceMath\cite{liu2025acemathadvancingfrontiermath}), progressed to using cutting-edge models (e.g., OpenAI (O1)\cite{gpt-o1}), and synthesized data via domain-specialized models for scalable knowledge transfer (e.g., DeepSeek-V3 \cite{deepseekai2024deepseekv3technicalreport}). This demonstrates a clear iterative path in data construction strategies. Currently, datasets used for reasoning enhancement are generally at the million-sample scale (e.g., Qwen-2.5-MATH \cite{yang2024qwen25mathtechnicalreportmathematical}). ii) \textit{Efficiency}. Adoption of "less is more" alignment (e.g., LIMA’s\cite{zhou2023limaalignment} 1k curated samples for 65B Llama \cite{touvron2023llama}, S1-32B’s\cite{muennighoff2025s1simpletesttimescaling} elite 1k high-diversity data comparable with O1-preview), proving minimal high-quality data optimally activates pretrained capabilities while reducing dependency on data scale. 

(2) \textit{Optimization Framework}. i) \textit{Sampling Strategies}. Recent advancements have seen a shift toward online reinforcement learning (RL), with approaches such as DeepSeek-V3 \cite{deepseekai2024deepseekv3technicalreport} and Qwen-2.5-MATH’s \cite{yang2024qwen25mathtechnicalreportmathematical} online sampling methods, which effectively mitigate distributional shifts. Additionally, Mini-Max employs an offline+online sampling strategy to enhance model performance. ii) \textit{Training Paradigms}. Multi-stage, collaborative optimization has become the dominant approach. For example, Llama 3 incorporates a six-round DPO iteration, while DeepSeek optimizes reasoning depth (long-CoT) and conciseness through temperature-varied sampling and reflection/verification prompts.
iii) \textit{Algorithms}. RL algorithms have evolved from early policy gradient methods to more complex PPO. Recent improvements based on PPO follow two main directions: one is to remove the critic model and train the policy using sparse rewards, thereby reducing the parameter count by half, as seen in DPO and GRPO; the other is to refine the design of the critic, such as PRIME \cite{cui2025processreinforcementimplicitrewards}, which introduces a ratio as the advantage function, and OREAL\cite{lyu2025exploringlimitoutcomereward}, which reshapes the rewards of positive and negative examples.

By prioritizing high-quality data and innovative optimization frameworks, the field is moving towards more effective and scalable models that can also better unlock the reasoning potential of MLLMs.

\subsection{Insight from LLM Alignment } 
Aligning LLMs has been a critical focus in recent research, offering valuable insights that can inform the development of MLLMs. By examining lessons learned from existing LLM alignment strategies, we can uncover key principles that may enhance MLLM community.

(1) \textit{Improving Training Efficiency}. Current MLLM alignment methods rely on the DPO loss function. However, since DPO requires loading both the policy model and reference model simultaneously, the training speed is significantly reduced. Could reference-free approaches like SimPO \cite{SimPO} be leveraged to further enhance training efficiency? This approach might accelerate the training process while reducing dependence on reference models. Further investigation into the specific role and impact of reference models in MLLM alignment is critical, as it could inform both efficiency improvements and optimized model design.

(2) \textit{Mitigating Overoptimization/Reward Hacking}. In LLM alignment using DPO~\cite{rafailov2024directpreferenceoptimizationlanguage} or RLHF, overoptimization remains a key challenge \cite{GaoSH23,Rafaelscaling}, where performance measured by the learned proxy reward model improves while true quality stagnates or deteriorates. Specifically, when training data exhibits significant quality disparities (e.g., excessive bias toward specific task types), models may overfit these task-specific patterns and underperform in real-world scenarios. Mitigation strategies include: i) using balanced training datasets to ensure diversity and representativeness, preventing narrow optimization; ii) implementing early stopping when validation performance plateaus; iii) incorporating regularization techniques to reduce overreliance on training data and enhance generalization.

\subsection{MLLM as Agents } 

MLLMs combine the powerful reasoning abilities of LLMs with the capability to understand and process data from multiple modalities, including images, texts, and audio. This enables them to extract knowledge from various types of information sources and perform integrated analysis, making them highly advantageous in handling complex real-world tasks \cite{xi2023risepotentiallargelanguage,wang2024mobileagentautonomousmultimodalmobile,ma2024surveyvisionlanguageactionmodelsembodied,durante2024agentaisurveyinghorizons,ma2024task}. First, the multimodal understanding capability of MLLMs provides a solid foundation for their application in complex tasks. For example, in the field of autonomous driving, MLLMs can process data from various modalities such as camera images, vehicle sensors, and traffic signals, allowing them to accurately perceive the surrounding environment \cite{wang2023drivemlmaligningmultimodallarge,cui2023surveymultimodallargelanguage}. Second, MLLMs inherit the powerful reasoning abilities of LLMs \cite{deepseekai2025deepseekr1incentivizingreasoningcapability}, enabling them to make precise decisions after perceiving the environment. When faced with complex tasks, MLLMs can break down the task and develop detailed execution plans to solve the problem efficiently. For instance, in industrial robotics, MLLMs can plan precise robot action sequences based on product assembly requirements, effectively directing the robot to complete various operations \cite{li2023manipllmembodiedmultimodallarge,liu2024selfcorrectedmultimodallargelanguage}. Furthermore, thanks to their powerful multimodal perception capabilities and the generalization abilities of LLMs across different domains, MLLMs exhibit excellent scalability and versatility, allowing them to be easily transferred to other tasks without the need for task-specific parameter adjustments.

However, there are still several pending issues that need to be addressed in order to transform MLLMs into highly effective agents. i) \textit{Multi-agent collaboration}. Current frameworks for multi-agent collaboration primarily focus on text-based agents, which have been shown to significantly expand the cognitive boundaries of individual agents. However, MLLM-based multi-agent systems still lack mature solutions, evident in the absence of tailored multimodal communication and information-sharing mechanisms \cite{ossowski2025commacommunicativemultimodalmultiagent}, as well as customized multimodal memory mechanisms. ii) \textit{Robustness}. The robustness of MLLM-based agents in open environments has not been systematically validated. For example, \cite{wu2025dissectingadversarialrobustnessmultimodal} demonstrated that adding adversarial perturbations to images in a web navigation environment can significantly hijack agents, forcing them to execute targeted adversarial goals. Subsequent research should introduce adversarial robustness testing and safeguarding techniques for MLLM agents. iii) \textit{Security}. The introduction of more complex components in MLLM agents increases the diversity of security risks. These risks span from multimodal environmental perception to reasoning and memory, with the potential for malicious attacks \cite{yang2024securitymatrixmultimodalagents}, leading to privacy breaches and hijacking. Future work should comprehensively explore diverse security protection mechanisms for MLLM agents to mitigate these risks.

%% file: draft/060.conclusion.tex
\section{Conclusion}

The field of MLLM alignment is developing rapidly.  In this paper, we conduct a systematic and comprehensive survey of existing research on MLLM alignment, focusing on four crucial questions:  what application scenarios can be covered, how to construct datasets, how to evaluate algorithms, and where the direction of the next alignment algorithm lies. As far as we know, this paper is the first systematic survey dedicated to MLLM alignment.  We hope that this survey will facilitate further research in this area.